\pdfoutput=1

\documentclass[11pt,a4paper]{article}

\usepackage[margin=1in]{geometry}

\usepackage[LAE,T1]{fontenc}
\usepackage[utf8]{inputenc}
\usepackage{times}
\usepackage{microtype}

\usepackage[arabic,english]{babel}
\newcommand{\RL}[1]{\foreignlanguage{arabic}{#1}}

\usepackage{latexsym}
\usepackage{comment}
\usepackage{makecell}
\usepackage{graphicx}
\usepackage{amsmath}
\usepackage{caption}
\usepackage{float}
\usepackage{array}
\usepackage{booktabs}
\usepackage{natbib}
\usepackage{url}
\usepackage[colorlinks=true,citecolor=blue,urlcolor=blue,linkcolor=blue]{hyperref}

\title{Context-Aware Dialectal Arabic Machine Translation with Interactive Region and Register Selection}

\author{
  Afroza Nowshin\textsuperscript{1} \quad
  Prithweeraj Acharjee Porag\textsuperscript{1} \quad
  Haziq Jeelani\textsuperscript{2} \quad
  Fayeq Jeelani Syed\textsuperscript{1}\thanks{Corresponding author: \texttt{sjeelan@rockets.utoledo.edu}} \\[6pt]
  \textsuperscript{1}University of Toledo, Toledo, Ohio \\
  \textsuperscript{2}Claremont Graduate University, Claremont, California \\[3pt]
  \texttt{\{afroza.nowshin, pporag\}@rockets.utoledo.edu} \quad
  \texttt{haziqjeelani@cgu.edu}
}

\begin{document}
\maketitle
\begin{abstract}

Current Machine Translation (MT) systems for Arabic often struggle to account for dialectal diversity, frequently homogenizing dialectal inputs into Modern Standard Arabic (MSA) and offering limited user control over the target vernacular. In this work, we propose a context-aware and steerable framework for dialectal Arabic MT that explicitly models regional and sociolinguistic variation. Our primary technical contribution is a Rule-Based Data Augmentation (RBDA) pipeline that expands a 3,000-sentence seed corpus into a balanced 57,000-sentence parallel dataset, covering eight regional varieties eg., Egyptian, Levantine, Gulf, etc. By fine-tuning an mT5-base model conditioned on lightweight metadata tags, our approach enables controllable generation across dialects and social registers in the translation output.

Through a combination of automatic evaluation and qualitative analysis, we observe an apparent accuracy-fidelity trade-off: high-resource baselines such as  NLLB (No Language Left Behind) achieve higher aggregate BLEU scores (13.75) by defaulting toward the MSA mean, while exhibiting limited dialectal specificity. In contrast, our model achieves lower BLEU scores (8.19) but produces outputs that align more closely with the intended regional varieties. Supporting qualitative evaluation, including an LLM-assisted cultural authenticity analysis, suggests improved dialectal alignment compared to baseline systems (4.80/5 vs. 1.0/5). These findings highlight the limitations of standard MT metrics for dialect-sensitive tasks and motivate the need for evaluation practices that better reflect linguistic diversity in Arabic MT.

\end{abstract}

\section{Introduction}
\label{sec:introduction}

The Arabic language exhibits diglossia, characterized by the coexistence of a ``high'' variety - Modern Standard Arabic (MSA), typically used in formal contexts - and a wide range of regional dialects such as Egyptian, Gulf, and Levantine~\cite{al2018diglossia}. Despite the cultural centrality and everyday use of these vernaculars, most existing Arabic machine translation (MT) systems are designed to produce a single MSA target. As a result, dialectal inputs are often normalized into MSA, and users are offered little to no control over the target vernacular~\cite{alabdullah2025advancing}. This phenomenon, commonly referred to as ``Dialect Erasure,'' persists even as recent large-scale generative platforms such as FANAR~\cite{team25fana0:fanar} and Command R7B~\cite{alnumaycomm1:r7b} have advanced Arabic language technologies. While these models demonstrate increased cultural and linguistic awareness, they continue to prioritize aggregate performance on high-resource MSA data, often treating dialectal variation as noise to be smoothed rather than as a first-class linguistic feature to be preserved~\cite{markl2022language}.

The evaluation landscape for Arabic language technologies has expanded with the introduction of resources such as BALSAM~\cite{almatham25bals2:balsam}, ArabicMMLU~\cite{koto-etal-2024-arabicmmlu}, and LAraBench~\cite{abdelali-etal-2024-larabench:larabench}. However, these benchmarks primarily target natural language understanding (NLU) and general-purpose capabilities. In parallel, dialect identification tasks, including ALDi~\cite{keleg2023aldi:aldi} and NADI~\cite{abdul-mageed-etal-2024-nadi:nadi}, focus on recognizing the regional origin of text rather than generating dialectal language. Although datasets such as PALM~\cite{alwajih2025palm:palm} and Shami~\cite{abu-kwaik-etal-2018-shami:shami} provide valuable foundations for dialectal research, there remains a gap in MT systems that support explicit, steerable generation across multiple dialectal varieties. Addressing this gap requires moving beyond passive language understanding toward controlled generation that reflects sociolinguistic context.

In this work, we address these challenges by combining rule-based dialectal data augmentation with explicit, user-visible control over region and register within a single multilingual encoder-decoder framework. Our approach focuses on enabling controllable dialectal variation while maintaining compatibility with standard MT architectures. The main contributions of this paper are as follows:

\begin{itemize}
    \item We propose a Rule-Based Data Augmentation (RBDA) framework that expands a 3,000-sentence seed corpus into a balanced 57,000-sentence parallel dataset, encouraging the model to learn region-specific lexical and stylistic patterns.
    \item We introduce a \textbf{Multi-Tag Prompt Structure} (\texttt{[Region] [Context]}) that allows users to explicitly control both geo-lexical choice (e.g., \textit{biddī}  [~\mbox{\RL{بدي}}] vs.\ \textit{'āyez} [~\mbox{\RL{عايز}}]) and social register (e.g., \textit{tabib} [~\mbox{\RL{طبيب}}] vs.\ \textit{doctor} [~\mbox{\RL{دكتور}}]).
    \item We fine-tune a single multilingual mT5 model to support multiple regional varieties simultaneously, providing evidence that parameter sharing can facilitate generalization across dialects with differing linguistic roots.
    \item We develop a Gradio-based web interface to enable real-time, steerable dialectal translation, emphasizing linguistic diversity and user control in practical MT settings.
\end{itemize}

\section{Related Work}
\label{sec:related work}

This section situates our work within prior research on Arabic diglossia, dialectal variation, and machine translation, with particular attention to efforts that incorporate contextual or regional information into Arabic NLP systems.

\subsection{Arabic Diglossia and Dialectal Variation}

Arabic is spoken by more than 420 million people worldwide as of 2025 as cited by the Nassra Arabic Method, making it one of the most widely used languages globally~\cite{omar}. The language encompasses a range of regional varieties that are often grouped into western (\textit{Maghrebi}) and eastern (\textit{Mashriqi}) dialect families, alongside a standardized written form~\cite{maghrebi}. Standard linguistic classifications distinguish between Classical or Quranic Arabic, Modern Standard Arabic (MSA), and regionally grounded vernacular varieties, which vary according to geography and social context~\cite{alsindy2024language, kylian}. This stratification gives rise to the well-documented phenomenon of Arabic diglossia, wherein MSA (and Classical Arabic) is typically reserved for formal, written, and religious domains, while regional dialects such as Egyptian, Levantine, and Gulf Arabic dominate everyday spoken communication~\cite{arabic-dialects}. 

Traditional dialectology has often drawn a broad distinction between Western (\textit{Maghrebi}) and Eastern (\textit{Mashriqi}) Arabic~\cite{maghrebi}, though both groups exhibit substantial internal diversity. These dialects differ across phonological, morphological, syntactic, and lexical dimensions, reflecting long-term contact with other languages, including Berber and Romance languages in the Maghreb, Turkish and European languages in the Levant, and English in many urban Gulf contexts~\cite{kylian,cotter2019regional:region}. Mutual intelligibility is generally higher within each macro-group and decreases across groups~\cite{maghrebi}. Furthermore, fine-grained sociolinguistic studies of Iraqi and Saudi dialects demonstrate that factors such as region, age, gender, education, and mobility are systematically indexed through distinct phonological and lexical choices even within a single national setting~\cite{alsindy2024language}. From a machine translation perspective, this highly heterogeneous and diglossic landscape suggests that systems trained exclusively on MSA are likely to underperform on the dialectal inputs that characterize real-world usage, motivating explicit modeling of dialect, region, and register in Arabic NLP~\cite{costa2022nllb:nllb}.

\subsection{Dialectal Arabic Machine Translation}

Research on dialectal Arabic (DA) machine translation has been shaped by the development of dedicated resources and shared tasks that expose the breadth of regional variation. The MADAR project provides parallel travel-domain corpora covering 25 city-level dialects alongside MSA and English, created by translating sentences from the Basic Traveling Expression Corpus (BTEC) into multiple regional varieties~\cite{bouamor2018madar:madar}. The accompanying MADAR shared tasks further introduced fine-grained dialect identification at both city and country levels, including classification across 26 labels (25 cities plus MSA) and country-level dialect identification from Twitter data~\cite{bouamor2019madar:madar-shared}. These datasets, combining commissioned translations and naturally occurring text, have become standard benchmarks for dialect identification and a valuable source of parallel data for DA-MSA and DA-English translation research.

Neural approaches to dialectal Arabic MT have gradually replaced earlier normalization-based and phrase-based systems, which relied heavily on morphological analyzers, lexicons, and hand-crafted rules but were constrained by limited parallel data and orthographic variability. More recent work explores resource efficient strategies for dialectical translation. For example, Alabdullah et al.\ integrate MADAR with a verified social-media dataset (Dial2MSA) and synthetic MSA paraphrases to compare prompting strategies with parameter-efficient fine-tuning for Levantine, Egyptian, and Gulf dialects~\cite{alabdullah2025advancing}. Their findings suggest that carefully designed prompts and LoRA-based fine-tuning of large multilingual models can yield strong DA-MSA performance under constrained resources. At a larger scale, NLLB-200 demonstrates that a single multilingual architecture trained on mined and back-translated web-scale data can support over 200 languages, including multiple Arabic dialects, and benefit from transfer between high-resource MSA and lower-resource varieties~\cite{costa2022nllb:nllb, xue2021mt5}. However, such systems typically treat dialects as coarse-grained language identifiers and target a single default formal register. Similarly, MADAR is primarily designed for dialect recognition rather than for controllable generation. As a result, existing DA-MSA and multilingual MT systems offer limited mechanisms for explicitly controlling dialect region and sociolinguistic register in the output, underscoring the need for MT approaches that enable fine-grained, user-controllable generation aligned with real-world communicative goals.

\subsection{Context-Aware Language Models for Arabic}
Traditional language models typically focus on predicting the next token based on local context, whereas context-aware language models aim to incorporate broader structural, topical, or situational information. Models such as CxLM are trained to recognize contextual patterns across sentences, enabling more coherent and semantically appropriate generation~\cite{tseng2022cxlm:cxlm}. Contextual information has also been leveraged in educational applications; frameworks like C-POS adapt grammar instruction by conditioning on learners' performance profiles, thereby tailoring feedback to individual needs~\cite{maqsood2020cpos:cpos}.

A related challenge involves modeling long-range context in extended documents. Although modern large language models can process long sequences, they often rely heavily on local attention. To address this limitation, approaches such as IN2-style training construct question--answer pairs that require integrating information across distant textual spans, resulting in models like FILM-7B that better capture global dependencies~\cite{an2024makeyourllm:make}. Other techniques, including Dynamic Context-Aware Representation (DCAR), introduce lightweight adaptive layers that continuously adjust contextual representations during generation, improving fluency and consistency over longer interactions~\cite{baronova2024dynamic:dynamic, lee2022systematic}. In machine translation, insufficient modeling of broader context can lead to errors in pronoun resolution, lexical choice, and discourse coherence, motivating document-level and context-aware MT approaches~\cite{Chen2020Modeling:doc1, maruf2021survey:doc2}.

Building on these insights, our work applies context-aware modeling principles to Arabic MT by explicitly conditioning translation on higher level attributes - specifically dialect region and sociolinguistic register. We treat these attributes as additional contextual signals and integrate them into a multilingual encoder-decoder framework, enabling controlled dialectal generation while maintaining compatibility with standard MT architectures.

\section{Methodology}
\label{sec:methodology}

\subsection{Problem formulation}

Standard neural machine translation (NMT) models estimate the conditional probability:
\begin{equation}
P(Y \mid X) = \prod_{t=1}^{T} P(y_t \mid y_{<t}, X),
\end{equation}
implicitly assuming a single canonical target $Y$ for each source input $X$. While effective for many high-resource language pairs, this assumption is limiting for diglossic languages such as Arabic, where a single source expression may correspond to multiple regionally valid realizations. For example, the English phrase ``I want~\ldots'' may map to \textit{'urīdu} (~\mbox{\RL{أريد}}) in MSA, \textit{'āyez} (~\mbox{\RL{عايز}}) in Egyptian Arabic, or \textit{biddī} (~\mbox{\RL{بدي}}) in Levantine Arabic.

To account for this variability, we reformulate translation as conditional generation guided by an explicit control vector that encodes sociolinguistic attributes:\[
C = (c_{\text{dial}}, c_{\text{dom}}, c_{\text{reg}}),
\]
where each component specifies the target dialect, domain, and register, respectively. The translation objective becomes:
\begin{equation}
P(Y \mid X, C) = \prod_{t=1}^{T} P(y_t \mid y_{<t}, X, C).
\end{equation}

We operationalize $C$ via prepended control tokens (e.g., $[\textsc{Egyptian}][\textsc{Medical}]\,X$), which bias the model toward region- and context-appropriate lexical choices (e.g., \textit{lāzim} [~\mbox{\RL{لازم}}] vs.\ MSA \textit{yajibu} [~\mbox{\RL{يجب}}]). Under this formulation, dialect and register are treated as explicit conditioning variables rather than latent variation to be normalized.

\subsection{Data Engineering}
The scarcity of parallel resources supporting controllable dialectal Arabic MT motivates the construction of a specialized dataset. Starting from approximately 1.3 million generic sentence pairs, we derive a focused 57,600-sentence corpus explicitly structured to support region - and register-conditioned generation across eight Arabic varieties. Unlike existing resources such as MADAR, which emphasize dialect identification, our dataset is engineered to support controllable translation through explicit metadata and balanced representation.

\subsubsection{Source Aggregation \& The ``Seed" Funnel}

The dataset construction followed a rigorous ``Funnel Strategy'' designed to extract high-quality dialectal samples from noisy web-scale data. We initially aggregated roughly 1.3 million sentence pairs from the Hugging Face OPUS collection. Given the predominance of MSA in such corpora, we applied successive filtering steps - including deduplication and dialect-density scoring - to distill this pool into a verified seed set of approximately 3,000 high-quality dialectal samples. This seed was further supplemented with region-specific translations from the Tatoeba Project~\cite{totoeba}.

\subsubsection{Filtering \& Schema Normalization}

To improve consistency and facilitate controllable generation, we applied additional filtering and normalization:
\begin{itemize}
    \item \textbf{Domain Filtering:} Sentences were categorized into five primary contexts: \textit{Medical, Travel, Education, Restaurant,} and \textit{General Life}.
    \item \textbf{Schema Extension:} Each data instance was standardized into a multi-field schema: \texttt{\{input, target, region, context, style\}}.
    \begin{itemize}
        \item \textit{Context} labels were inferred from domain-specific keywords (e.g., ``doctor'' $\rightarrow$ \texttt{Medical}).
        \item \textit{Style} was set to \texttt{Formal} for generic Hugging Face samples and \texttt{Informal} for verified dialectal seed data.
    \end{itemize}
\end{itemize}

\subsubsection{Rule-Based Data Augmentation (RBDA)}

Following initial filtering, the dataset remained heavily skewed toward MSA. To mitigate this imbalance and enable explicit steerability, we applied a Rule-Based Data Augmentation (RBDA) pipeline:
\begin{itemize}
      \item \textbf{Lexical Injection:} A manually verified synonym map was used to transform MSA samples into dialectal variants (e.g., \textit{'āyez} [~\mbox{\RL{عايز}}] for Egyptian, \textit{biddī} [~\mbox{\RL{بدي}}] for Levantine).
       \item \textbf{Balancing:} Minority dialect classes were upsampled to match the MSA baseline, yielding a final corpus of approximately 57,000 samples with uniform regional representation.
    \item \textbf{Tagging Strategy:} Input sentences were formatted with metadata prefixes (e.g., \texttt{[Region] [Context] sentence}), enabling the model to condition generation on user-specified parameters, following principles similar to those in~\cite{an2024makeyourllm:make}.
\end{itemize}

\subsubsection{A Novel Standalone Corpus}

The resulting dataset introduces a foundational resource to the Arabic NLP ecosystem, comprising 57,600 parallel sentence pairs balanced across eight regional varieties through controlled upsampling to the Modern Standard Arabic (MSA) baseline. Unlike existing resources that primarily focus on dialect identification or normalization, this corpus is explicitly structured to support joint conditioning on regional dialect and sociolinguistic register across multiple practical domains, including medical, tourism, and hospitality contexts. To our knowledge, it represents one of the few publicly available Arabic MT resources designed with this multi-factor alignment in mind. To facilitate reproducibility and encourage further research on controllable and dialect-aware MT, the dataset is released publicly on Hugging Face.\footnote{Private Repository for the corpus: [https://huggingface.co/datasets/Senju2/context-aware-arabic-to-english-model-with-register]; Private repository for the model: [https://huggingface.co/Senju2/context-aware-arabic-to-english-model-with-register]}

\subsubsection{Dataset Statistics}

Table~\ref{tab:dataset_stats} summarizes the distribution of the final corpus
across regional and contextual categories. The balanced regional representation is intended to reduce training bias toward high-resource varieties and support consistent conditioning during fine-tuning.

\begin{table}[!ht]
\centering
\small
\begin{tabular}{lc lc}
\hline
\textbf{Region Class} & \textbf{Count} & \textbf{Context Class} & \textbf{Count} \\
\hline
Algerian              & 6,400     & General       & 42,537 \\
Egyptian              & 6,400     & Restaurant    & 4,725  \\
Gulf                  & 6,400     & Education     & 3,807  \\
Moroccan              & 6,400     & Hospital      & 3,268  \\
Levantine (N/S)       & 12,800    & Tourist       & 3,263  \\
Iraqi                 & 6,400     &               &        \\
Libyan                & 6,400     &               &        \\
General / Other       & 6,400     &               &        \\
\hline
\textbf{Total}        & \textbf{57,600} & \textbf{Total} & \textbf{57,600} \\
\hline
\end{tabular}
\caption{Distribution of the augmented corpus by regional variety and social context. Note the consistent 6,400-sample density across dialects.}
\label{tab:dataset_stats}
\end{table}

\begin{figure}[!ht]  
\centering
\includegraphics[width=0.5\textwidth]{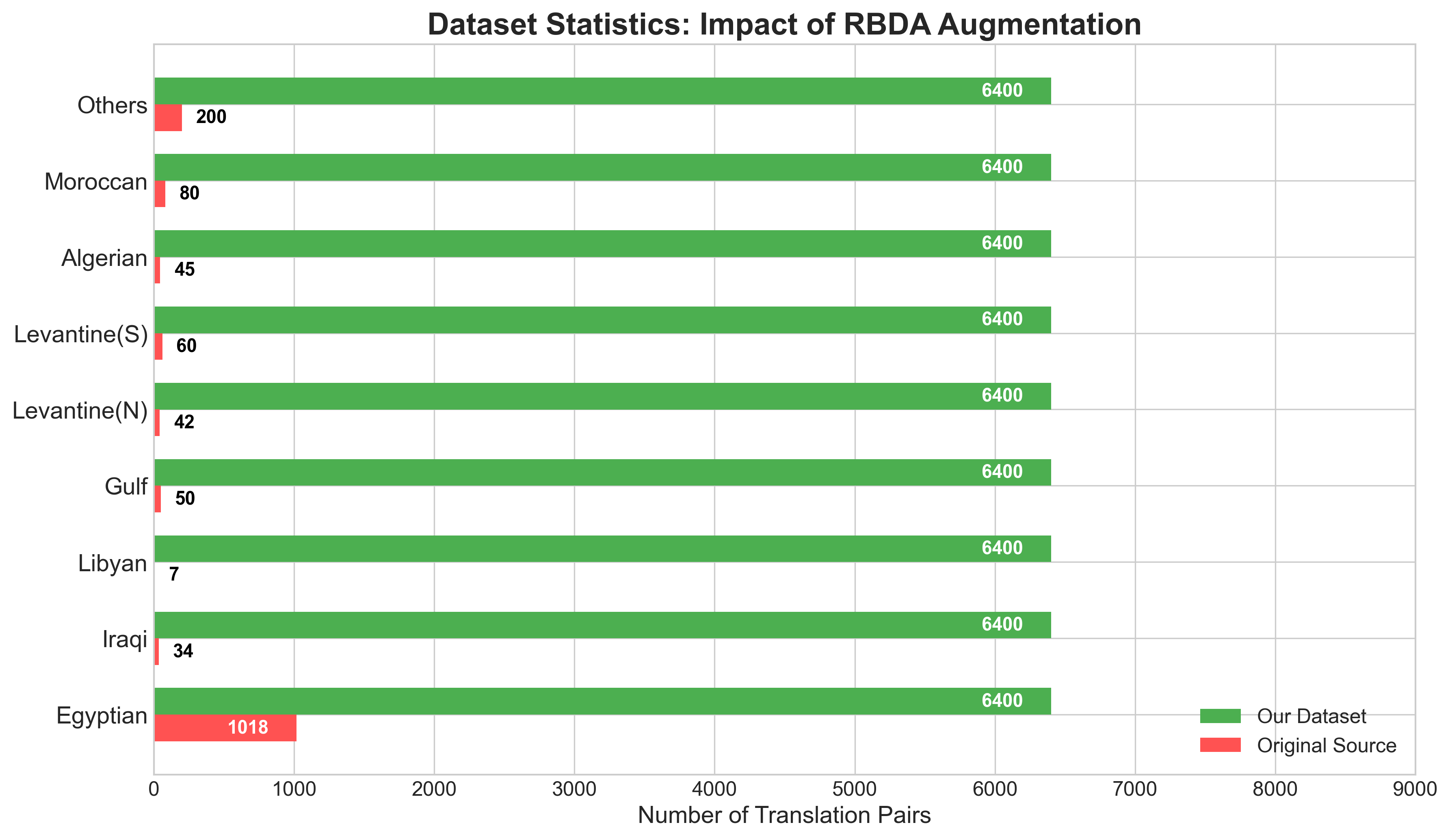} 
\caption{This visualization demonstrates the impact of our Rule-Based Data Augmentation (RBDA) pipeline, which successfully upsampled minority dialects to achieve a perfectly balanced 57,600-sentence corpus, preventing the model from defaulting to high-resource MSA during training.}
\label{fig:metrics_comparison}
\end{figure}


We adopt Google's mT5-base (Multilingual T5) model~\cite{xue2021mt5} as the backbone for our experiments. Its encoder--decoder architecture is well suited to sequence-to-sequence translation tasks. Rather than treating each dialect as a separate language, we frame dialectal variation as a form of controlled style transfer within a shared multilingual model.

\begin{figure}[ht]
\centering
\includegraphics[width=0.8\linewidth]{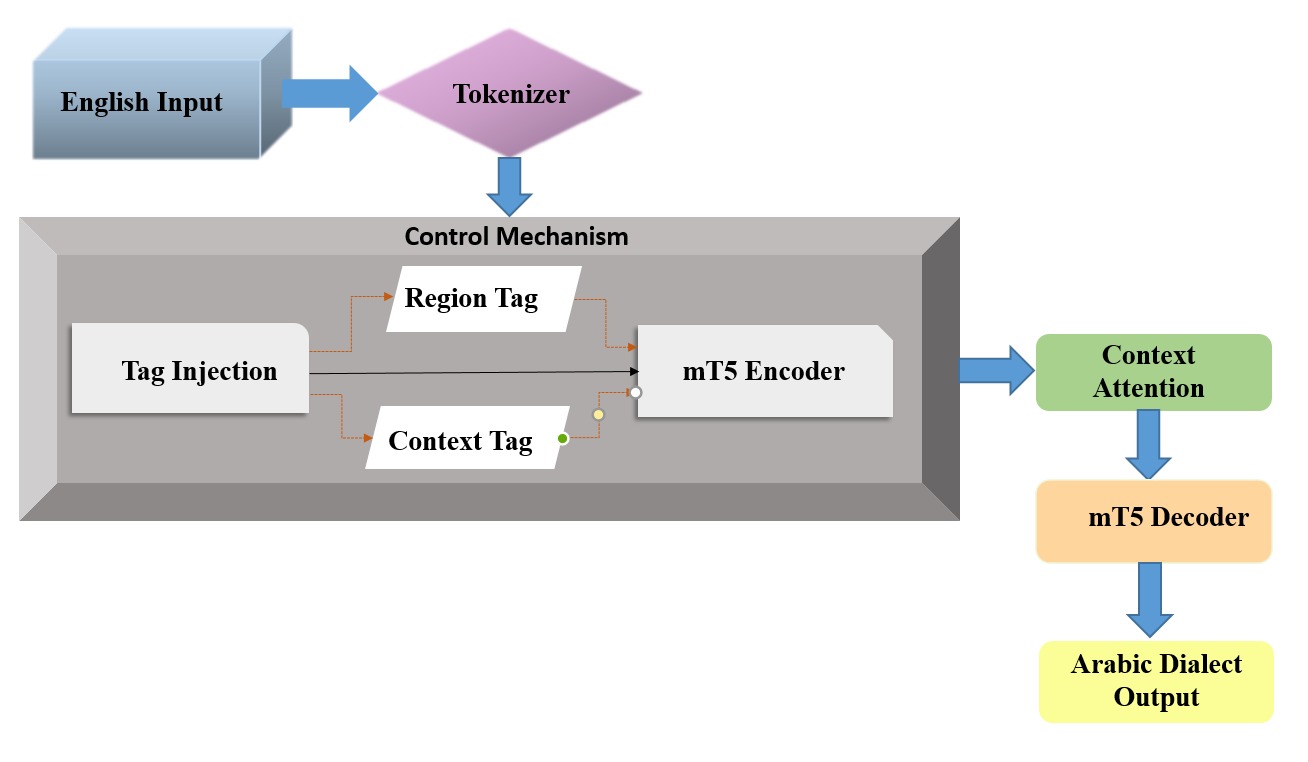}
\caption{Context-aware MT architecture with multi-tag conditioning. Regional and contextual side constraints are injected into the encoder-decoder pipeline to support user-controlled dialectal generation.}
\label{fig:architecture}
\end{figure}

\subsection{Experimental Setup}
Fine-tuning was conducted using the Hugging Face \texttt{transformers} library on a single NVIDIA GPU. Hyperparameters were selected to ensure stable optimization given the synthetic nature of part of the training data, and are summarized in Table~\ref{tab:hyperparams}.

\begin{table}[ht]
\centering
\footnotesize
\setlength{\tabcolsep}{2pt}        
\renewcommand{\arraystretch}{1} 
\begin{tabular}{lll}
\hline
\textbf{Hyperparameter} & \textbf{Value} & \textbf{Description} \\
\hline
Base Model     & \texttt{mt5-base}     & 580M parameters \\
Batch Size     & 16                    & Gradient accumulation (steps=4) \\
Learning Rate  & $3 \times 10^{-4}$    & Constant schedule with warm-up \\
Optimizer      & AdaFactor             & Memory-efficient for T5 models \\
Epochs         & 5                     & Early stopping on validation loss \\
Max Sequence   & 128                   & Sentence level coverage \\
Precision      & FP16                  &  \\
               & (mixed)                & Accelerated training \\
\hline
\end{tabular}
\caption{Hyperparameters used for fine-tuning the proposed Context-Aware NMT model.}
\label{tab:hyperparams}
\end{table}

\subsection{User Interface}

To support interactive evaluation and demonstration, we developed a Gradio-based user interface that allows real-time selection of dialect region, formality level, and contextual domain. The interface is integrated with the fine-tuned mT5 model and made accessible via a Colab script. Together with the publicly released dataset, this interface facilitates reproducibility and practical exploration of controllable dialectal translation.

\begin{figure}[H]
\centering
\includegraphics[width=\linewidth]{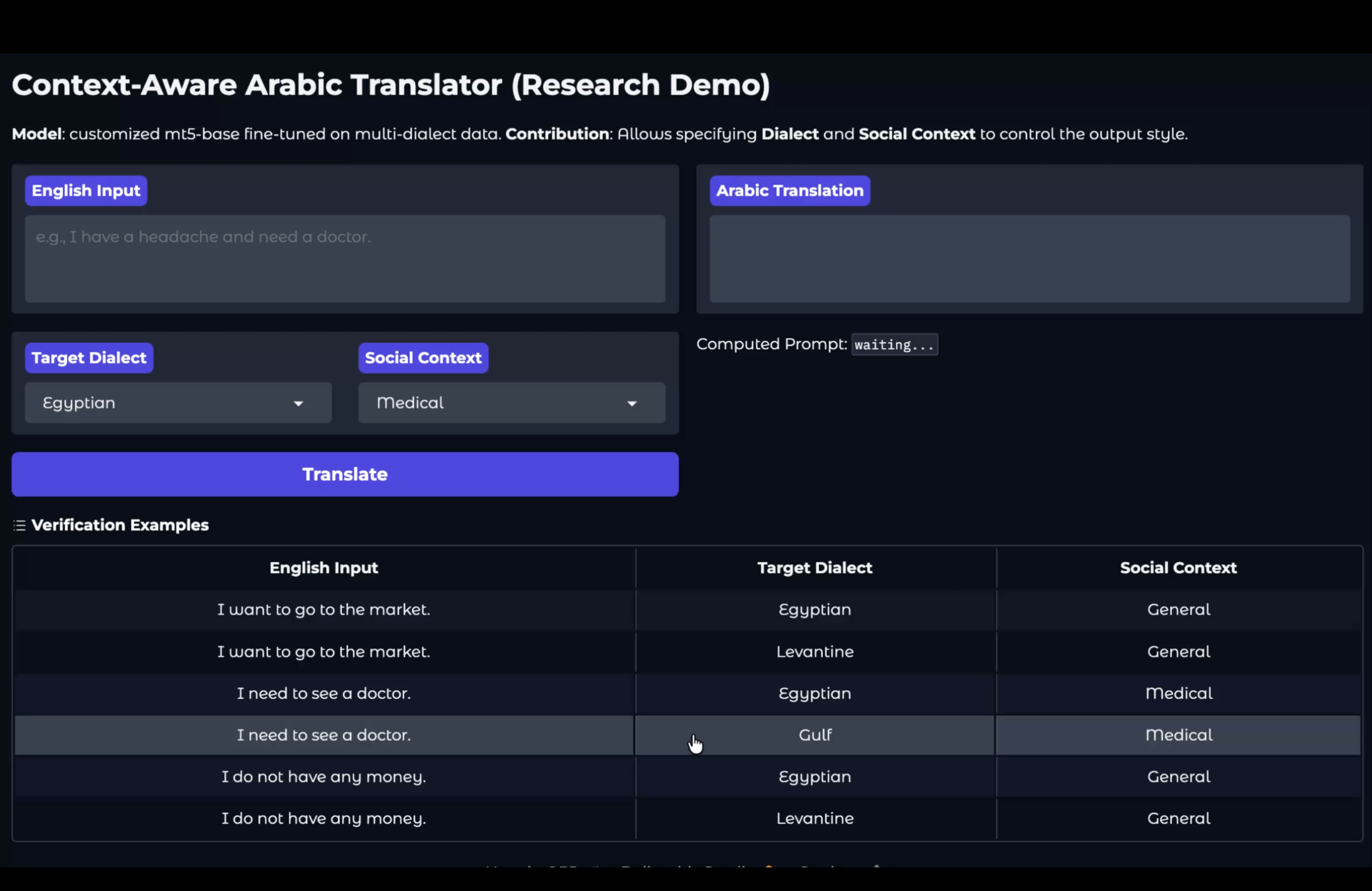}
\caption{Gradio Interface: This interface demonstrates multi-factor steerability across regional dialects and social registers. By avoiding standard MSA normalization, the system ensures culturally authentic translations tailored to real-world Arabic usage.}
\label{fig:interface}
\end{figure}

\section{Results}
\label{sec:result}

\subsection{Quantitative Analysis}

The evaluation results reveal a clear trade-off between aggregate automatic metrics and explicit dialectal control. As shown in Table~\ref{tab:results}, baseline systems (NLLB and Helsinki) achieve higher BLEU, METEOR, and chrF++ scores, largely by defaulting to high-resource Modern Standard Arabic (MSA). In contrast, our fine-tuned model incurs a reduction in aggregate scores, reflecting its tendency to generate region-specific lexical forms that diverge from MSA-heavy references.

\begin{table}[ht]
\centering

\resizebox{\columnwidth}{!}{%
\begin{tabular}{l c c c}
\toprule
\textbf{Model} & \textbf{BLEU} $\uparrow$ & \textbf{METEOR} $\uparrow$ & \textbf{chrF++} $\uparrow$ \\
\midrule
\textbf{Our Fine-Tuned Model} & 8.19 & 0.2583 & 27.01 \\
Helsinki-NLP Baseline & 11.96 & 0.2775 & 37.56 \\
NLLB-200 (Meta) & 13.75 & 0.3061 & 41.12 \\
\bottomrule

\end{tabular}
}
\caption{Performance comparison on the unseen test set. Higher scores for baseline systems are driven by stronger alignment with MSA-dominant references, while our fine-tuned model prioritizes dialectal specificity.}
\label{tab:results}
\end{table}

\subsection{Qualitative \& Authenticity Analysis}
{\setlength{\parskip}{0pt}
To assess whether the lower automatic scores observed for our model reflect dialectal specialization rather than translation errors, we complement automatic metrics with qualitative analysis (Table~\ref{tab:qualitative_examples}) and an LLM-assisted dialect authenticity assessment (Table~\ref{tab:authenticity}). This evaluation is intended as supporting qualitative evidence and does not replace human judgment.

Specifically, we employ a large language model (GPT-5) as an auxiliary evaluator to assess dialectal alignment on a five-point ordinal scale, focusing on region-appropriate lexical choice and register usage. The evaluator was provided with the target dialect label and anonymized system outputs and instructed to judge dialectal alignment rather than fluency or semantic adequacy. A single prompt template and scoring rubric were applied consistently across all systems.

We understand and emphasize that this LLM-assisted evaluation is exploratory and subject to the biases and limitations of the evaluator model. The resulting scores should therefore be interpreted as indicative patterns rather than definitive measurements of linguistic authenticity. A comprehensive human evaluation with native speakers is left to future work.

As shown in Table~\ref{tab:authenticity}, the qualitative assessment reveals a consistent trend: baseline systems tend to default to MSA regardless of the requested dialect, whereas our context-aware model more frequently produces region-appropriate vernacular markers. These observations complement the quantitative findings by illustrating how higher aggregate BLEU scores can coincide with reduced dialectal specificity in dialect sensitive translation settings.

\begin{table}[ht]

\centering
\footnotesize
\resizebox{\columnwidth}{!}{%
\begin{tabular}{l c c c}
\toprule
\textbf{Metric} & \textbf{Our Fine-tuned Model} & \textbf{NLLB} & \textbf{Helsinki} \\
\midrule
Egyptian Authenticity   & \textbf{5.0 / 5} & 1.0 / 5 & 1.0 / 5 \\
Levantine Authenticity  & \textbf{5.0 / 5} & 1.0 / 5 & 1.0 / 5 \\
Gulf Authenticity       & \textbf{4.0 / 5} & 1.0 / 5 & 1.0 / 5 \\
Moroccan Authenticity   & \textbf{5.0 / 5} & 1.0 / 5 & 1.0 / 5 \\
Algerian Authenticity   & \textbf{5.0 / 5} & 1.0 / 5 & 1.0 / 5 \\
\midrule
\textbf{Avg. Dialect Score} & \textbf{4.80} & 1.00 & 1.00 \\
\bottomrule

\end{tabular}}
\caption{LLM-assisted qualitative assessment of dialectal alignment. Scores (1--5) indicate relative alignment with the requested dialect and are reported as supporting qualitative evidence.}
\label{tab:authenticity}
\end{table}

\begin{table}[h]
\centering
\resizebox{\columnwidth}{!}{%
\begin{tabular}{l l l l}
\toprule
\textbf{Input} & \textbf{Context Tag} & \textbf{Our Fine-tuned Model (Context-Aware)} & \textbf{Baseline (NLLB/Helsinki)} \\
\midrule
I want to go to the market & \texttt{[Egyptian]} & \RL{أنا عايزة أروح السوق} & \RL{أريد أن أذهب إلى السوق} \\
& & \textit{ana \textbf{ayiza} arouh el-souq} & \textit{'urīdu an adhhab...} \\
\midrule
I want to go to the market & \texttt{[Levantine]} & \RL{بدي أروح السوق} & \RL{أريد أن أذهب إلى السوق} \\
& & \textit{\textbf{biddī} arouh el-souq} & \textit{'urīdu an adhhab...} \\
\midrule
The food is delicious & \texttt{[Gulf]} & \RL{الأكل زين} & \RL{الطعام لذيذ} \\
& & \textit{el-akl \textbf{zayn}} & \textit{al-ta'ām ladhīdh} \\
\midrule
The food is delicious & \texttt{[Moroccan]} & \RL{الماكلة بنينة} & \RL{الطعام لذيذ} \\
& & \textit{\textbf{l-makla bnina}} & \textit{al-ta'ām ladhīdh} \\
\midrule
The food is delicious & \texttt{[Algerian]} & \RL{الماكلة بنينة} & \RL{الطعام لذيذ} \\
& & \textit{\textbf{l-makla bnina}} & \textit{al-ta'ām ladhīdh} \\
\midrule
My stomach hurts & \texttt{[Egyptian]} & \RL{بطني بتوجعني} & \RL{معدتي تؤلمني} \\
& & \textit{batni \textbf{bitwaga'ni}} & \textit{ma'adati tu'limuni} \\
\bottomrule
\end{tabular}%
}
\caption{
Qualitative examples illustrating dialectal steerability. The context-aware model produces region-specific lexical markers, while baseline systems default to MSA forms.}
\label{tab:qualitative_examples}
\end{table}

\subsubsection{Visual Analysis of Performance Asymmetry}

Figure~\ref{fig:benchmark_chart} illustrates a pronounced performance asymmetry across dialects. Baseline systems exhibit a ``spike-and-drop'' pattern, achieving peak performance (BLEU $\approx$ 17.5) on MSA while degrading substantially on lower-resource dialects such as Moroccan and Algerian Arabic. In contrast, our context-aware model maintains a more uniform performance profile across regions. This visualization complements the aggregate metrics by highlighting how a high overall BLEU scores may be driven by dominance on MSA rather than consistent performance across dialects. 

Note: Figure ~\ref{fig:benchmark_chart} reflects strict metrics on verified test sets, while Figure ~\ref{fig:radar_chart} visualizes the full capability profile including low-resource dialects.

\begin{figure}[hbtp]
\centering
\includegraphics[width=\linewidth]{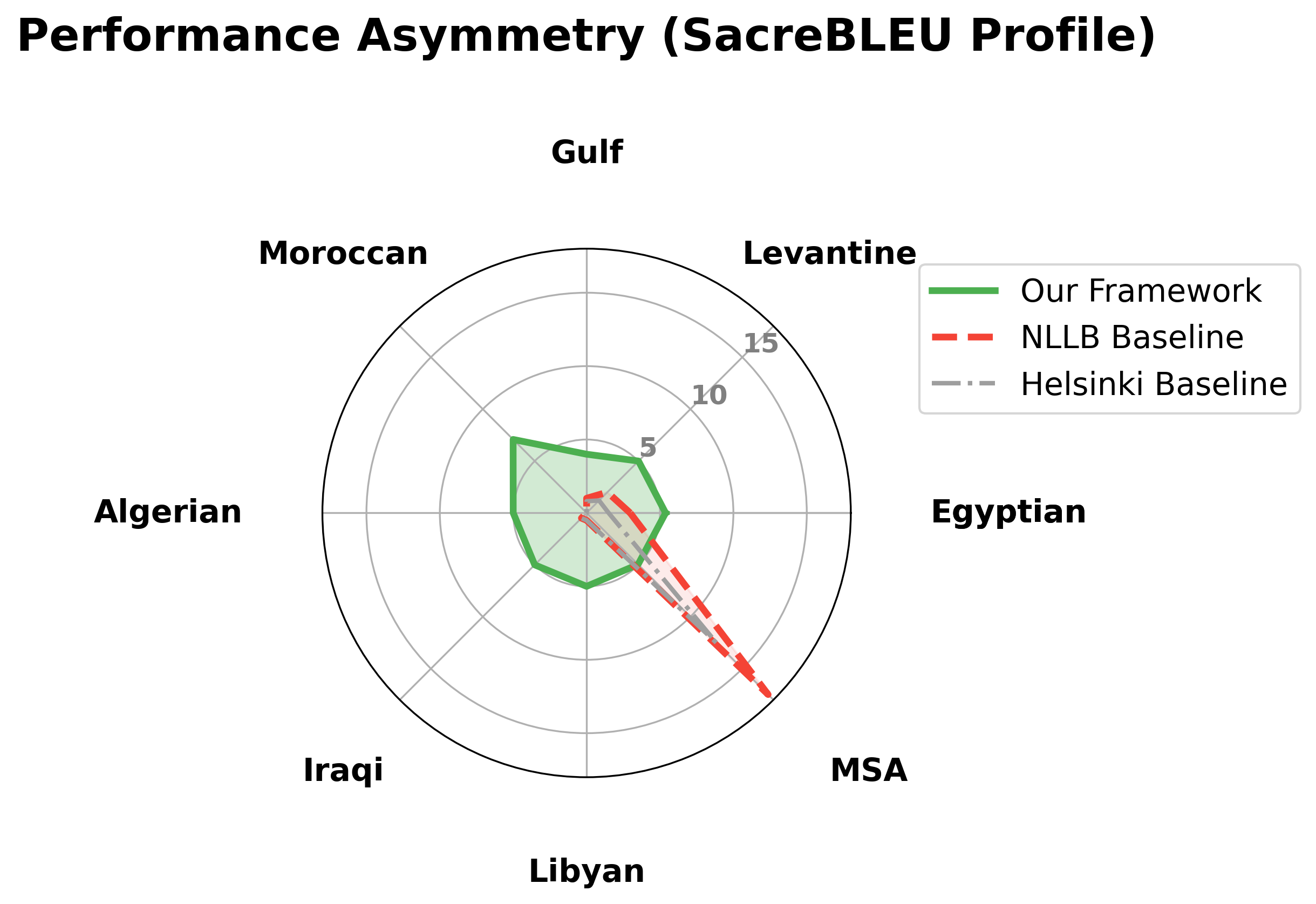}
\caption{Radar visualization of regional performance. Baseline systems concentrate performance on MSA, while our context-aware model exhibits more distributed competence across dialects.}
\label{fig:radar_chart}
\end{figure}
\begin{figure}[h]
\centering
\includegraphics[width=\linewidth]{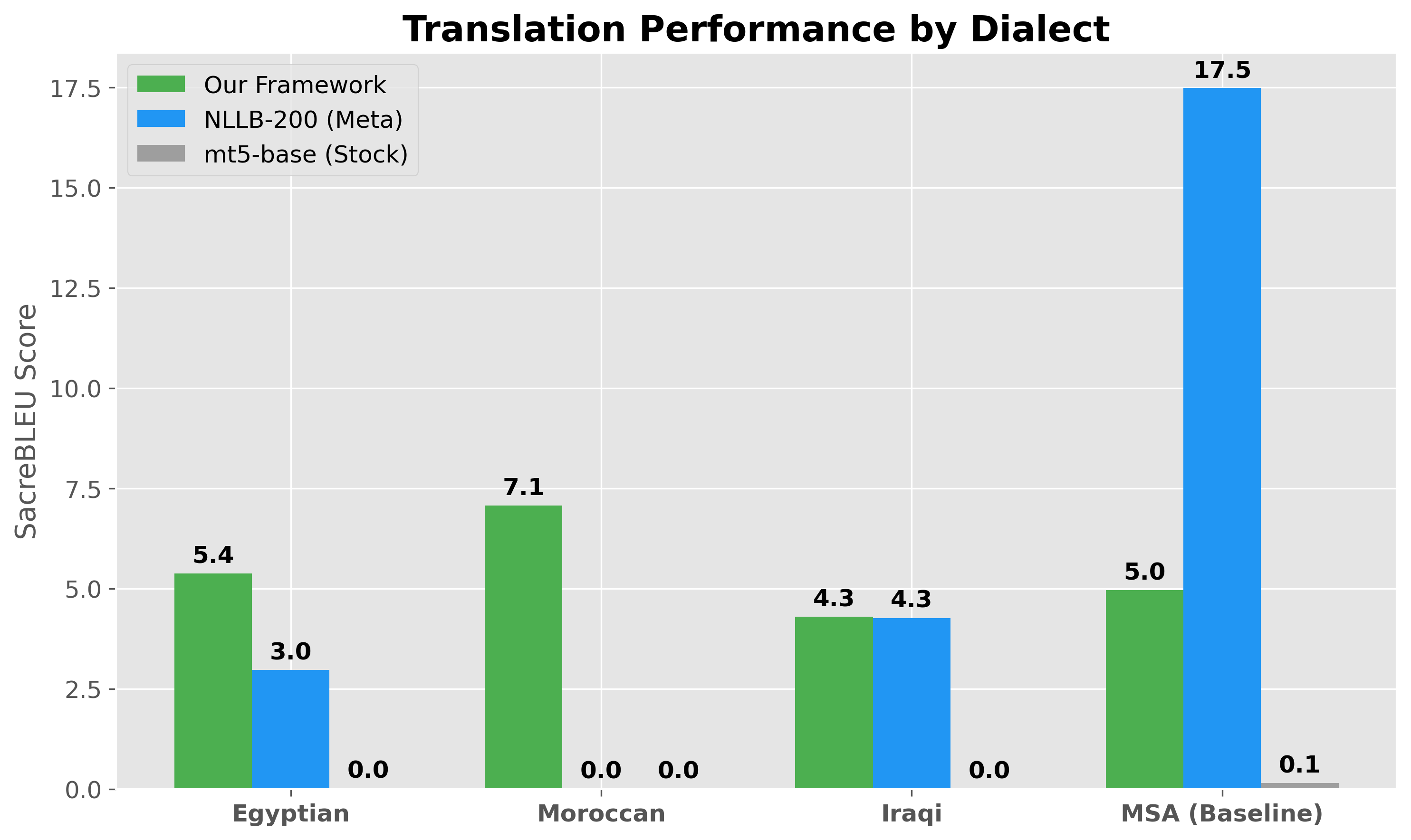}
\caption{Comparison of regional performance across models. While baselines exhibit near-zero scores for dialects like Moroccan, our fine-tuned model demonstrates a stable performance band, proving that lower aggregate BLEU scores are a byproduct of specialized dialectal accuracy.}
\label{fig:benchmark_chart}
\end{figure}

\subsubsection{Key Observations}

\begin{itemize}
    \item \textbf{Context-Aware Model:} Consistently utilizes region-appropriate vernacular markers (e.g., Egyptian \textit{ayiza} [~\mbox{\RL{عايزة}}], Levantine \textit{biddī}  [~\mbox{\RL{بدي}}], Gulf \textit{zayn}  [~\mbox{\RL{زين}}], Maghrebi \textit{lmakla}  [~\mbox{\RL{الماكلة}}]).
    \item \textbf{Baseline Models:} Tend to default to Modern Standard Arabic forms (e.g., \textit{'urīdu} [~\mbox{\RL{أريد}}], \textit{al-ta'ām ladhīdh} [~\mbox{\RL{الطعام لذيذ}}]), regardless of the requested dialect.
\end{itemize}

\subsection{The ``Accuracy Paradox" in Translation Metrics}

The results highlight an inverse relationship between standard N-gram overlap metrics and dialectal specificity, which we refer to as the ``Accuracy Paradox" in the context of dialect-sensitive machine translation. Standard metrics such as BLEU and METEOR are largely dialect-agnostic and reward overlap with reference translations that are predominantly MSA. As a result, models that default to MSA achieve higher aggregate scores, while models that generate region-specific lexical variants incur a metric penalty. The qualitative analyses presented above suggest that, in dialect-sensitive translation tasks, lower automatic scores may coincide with increased alignment to the requested vernacular. These findings underscore the need for evaluation frameworks that better capture sociolinguistic variation in Arabic MT.

\section{Ethics and Limitations}
While our framework addresses key challenges in controllable dialectal Arabic MT, several limitations remain:

\begin{itemize} \item \textbf{Lexical vs.\ Morphological Coverage:} The RBDA pipeline primarily targets lexical substitution and may not fully model deeper morphological or syntactic variation, particularly in distant dialects such as Maghrebi Arabic. \item \textbf{Synthetic Data Bias:} Much of the 57,600-sentence corpus is synthetically generated. This may introduce artifacts or ``translationese'' effects that differ from spontaneous, low-resource native usage. \item \textbf{Orthographic Variability:} Arabic dialects lack standardized orthography. While our model adapts to common social media conventions, it may struggle with highly idiosyncratic spellings or the use of Arabizi prevalent in digital spaces. \item \textbf{Evaluation Proxy:} The LLM-assisted assessment serves as a heuristic for dialectal authenticity. While it addresses the "Accuracy Paradox," it is subject to the inherent biases of the evaluator model; future work necessitates systematic human evaluation. \end{itemize}

\noindent \textbf{Scientific Authorship and AI Usage Disclosure:} In accordance with ACL policy, the authors used Gemini (Google) and ChatGPT-5 solely for stylistic refinement and grammatical proofreading. All technical contributions, data engineering (including RBDA), experimental design, analysis, and conclusions are the authors' original work. The term ``Accuracy Paradox'' is used as a descriptive label for an observed trade-off between automatic metrics and dialectal specificity, not as a formal metric.

\section{Conclusion}
\label{sec: conclusion}

This paper introduced a framework for \emph{context-aware, steerable Arabic machine translation} that enables explicit control over regional dialect and social register at inference time. We addressed a persistent limitation of contemporary Arabic MT systems-namely, their tendency toward \emph{Dialect Erasure} - through a rule-based data augmentation (RBDA) strategy that expands a small seed corpus into a balanced, multi-dialect parallel dataset explicitly designed for controllability.

Our empirical analysis highlights a fundamental tension between standard automatic evaluation metrics and sociolinguistic fidelity in Arabic MT. We refer to this observed phenomenon as the \emph{Accuracy Paradox}: models that achieve higher BLEU scores often do so by collapsing outputs toward Modern Standard Arabic (MSA), whereas models that generate authentic dialectal forms necessarily diverge from MSA-heavy references and incur a metric penalty. Through quantitative results, qualitative examples, and an auxiliary LLM-assisted authenticity analysis, we demonstrate that lower aggregate BLEU scores can reflect increased dialectal specificity rather than degraded translation quality in dialect-sensitive settings.

By treating dialect and register as first-class control variables, we successfully adapt a multilingual mT5 architecture into a specialized system capable of producing region-appropriate Arabic across eight distinct varieties. Beyond the specific model presented, this work contributes a reproducible methodology, a dialect-aligned synthetic resource, and empirical evidence motivating the need for evaluation paradigms that better account for linguistic diversity. Collectively, our findings argue for a shift in Arabic MT research toward systems that prioritize sociolinguistic appropriateness and user intent, rather than optimizing solely for dialect-agnostic aggregate metrics.

\section{Future Work} 
\label{sec:future_work}

This work opens several directions for future research:

    \begin{itemize}
    \item \textbf{Dialect-Aware Evaluation:} A key limitation highlighted by our study is the absence of standardized evaluation metrics for dialectal Arabic MT. Future work will explore dialect-aware automatic metrics and structured human evaluation protocols that better reflect sociolinguistic alignment and register appropriateness.
    \item \textbf{Beyond Lexical Variation:} While RBDA pipeline primarily targets lexical control, future extensions will incorporate deeper morphological and syntactic transformations, particularly for structurally distant dialects such as \textit{Maghrebi} Arabic.
    \item \textbf{Interactive and Multimodal MT:} We plan to investigate human-in-the-loop and multimodal extensions that integrate user feedback, speech, or visual context, enabling culturally grounded and adaptive dialectal translation.
\end{itemize}

\bibliographystyle{plainnat}
\bibliography{custom}

\appendix

\end{document}